\documentclass{article}

\usepackage{arxiv}

\usepackage[utf8]{inputenc} 
\usepackage[T1]{fontenc}    
\usepackage{hyperref}       
\usepackage{url}            
\usepackage{booktabs}       
\usepackage{amsfonts}       
\usepackage{nicefrac}       
\usepackage{microtype}      
\usepackage{lipsum}
\usepackage{graphicx}
\usepackage{multirow}

\title{SPARTA: Speaker Profiling for ARabic TAlk}

\begin{document}
\author{
  Wael Farhan\\
  \texttt{wael.farhan@mawdoo3.com} \\
  Mawdoo3 Ltd\\
  Amman, Jordan
  \And
  Muhy Eddin Za'ter \\
  \texttt{muhy.zater@mawdoo3.com} \\
    Mawdoo3 Ltd\\
  Amman, Jordan
  \AND
  Qusai Abu Obaidah\\
  \texttt{qusai.abuobaida@mawdoo3.com} \\
    Mawdoo3 Ltd\\
  Amman, Jordan
  \And
  Hisham al Bataineh \\
  \texttt{hisham.bataineh@mawdoo3.com} \\
    Mawdoo3 Ltd\\
  Amman, Jordan
  \And
  Zyad Sober \\
  \texttt{zyad.sober@mawdoo3.com} \\
    Mawdoo3 Ltd\\
  Amman, Jordan
  \AND
  Hussein T. Al-Natsheh\\
  \texttt{h.natsheh@mawdoo3.com} \\
  Mawdoo3 Ltd\\
  Amman, Jordan
}

\maketitle
\begin{abstract}
\label{sec:abstract}
This paper proposes a novel approach to an automatic estimation of three speaker traits from Arabic speech: gender, emotion, and dialect. After showing promising results on different text classification tasks, the multi-task learning (MTL) approach is used in this paper for Arabic speech classification tasks. The dataset was assembled from six publicly available datasets. First, The datasets were edited and thoroughly divided into train, development, and test sets (open to the public), and a benchmark was set for each task and dataset throughout the paper. Then, three different networks were explored: Long Short Term Memory (LSTM), Convolutional Neural Network (CNN), and Fully-Connected Neural Network (FCNN) on five different types of features: two raw features (MFCC and MEL) and three pre-trained vectors (i-vectors, d-vectors, and x-vectors). LSTM and CNN networks were implemented using raw features: MFCC and MEL, where FCNN was explored on the pre-trained vectors while varying the hyper-parameters of these networks to obtain the best results for each dataset and task. MTL was evaluated against the single task learning (STL) approach for the three tasks and six datasets, in which the MTL and pre-trained vectors almost constantly outperformed STL. All the data and pre-trained models used in this paper are available and can be acquired by the public. 
\end{abstract}


\section{Introduction}
\label{sec:intro}
Speech is the basic form of human-to-human communication. Its signals hold valuable information about the speaker's identity, as they can convey the speaker's age, gender, emotional state, dialect, and other characteristics. Therefore, automatic detection of the speaker's features can inevitably enhance and affect various fields and has a wide range of applications, including, but not limited to, medical, commercial, forensics, and other real-world applications that involve human-machine interaction scenarios. Thus, intensive research endeavors recently started focusing on estimating and extracting important traits of speakers solely from speech.

The work presented in this paper concentrates on classifying the speaker's gender, dialect, and emotional tone. In most human-machine interactions, it is necessary that the gender of the speaker or user to be known. Emotion, on the other hand, is a fundamental factor in any human-to-human communication. The speaker's voice pitch, intonation, and rate, they all affect the emotions perceived by the listener and hence can change the semantics of the utterance, which in turn makes installing an effective emotion recognition system in human-machine interaction of great benefit. Moreover, recognizing the dialect of the speaker simplifies speech-related tasks such as Automatic Speech Recognition (ASR), which makes the task of dialect identification an important part of speech recognition systems. The approach presented in this paper; Multi-task learning, has not been used to detect these traits of speakers from Arabic speech or any language in any previous work.

Although voice signals are mixed signals from which multiple features can be extracted simultaneously, until now, the majority of the voice-related recognition tasks pay more attention to the identification of a single feature from any voice utterance \cite{koolagudi2012emotion}. By contrast, few recognition systems can identify multiple features such as the speaker's gender and emotion simultaneously \cite{chen2018multi,poorjam2014multitask}. The availability of speech recognition systems drastically decreases in the case of the Arabic language in line with the deficiency of available datasets.

In this paper, a novel multi-task learning framework is proposed for the detection of gender, emotion, and dialect simultaneously from Arabic speech, while also exploring various feature extraction methods and speech utterance modeling techniques. The dataset used to train and test the models were assembled and collected from existing datasets that can be acquired by the public. The division of datasets and their results can be used as a benchmark for future work\footnote{https://github.com/mawdoo3/sparta-benchmark}. Finally, the method proposed in this paper outperforms the best result recorded in the literature on the tasks of one of the datasets.

The rest of the paper is structured as follows: section \ref{sec:related_work} presents related work in this area. Section \ref{sec:dataset} describes the collected datasets. Section \ref{sec:feature_extraction} demonstrates various feature extraction methods. The proposed method for learning multi-task classification is described in section \ref{sec:method}. Section \ref{sec:results_and_discussion} presents the results obtained from the conducted experiments. Finally, section \ref{sec:conclusion} concludes the work presented in the paper. 

\section{Related Work}
\label{sec:related_work}
The detection and estimation of the features and speaker's traits from speech have been a challenge for researchers and industries since the emergence of speech recognition \cite{radha2012review,desai2013feature} due to multiple factors, including, but not limited to: background noise, different accents, and speaker dependence. However, research endeavors were able to propose alternative solutions and techniques to tackle this challenge. 

\subsection{Speaker Profiling}
Researchers followed different approaches and explored various techniques for different speech recognition and detection tasks such as ASR, gender, emotion, and dialect detection \cite{ververidis2004automatic,palo2016efficient}. Previous work started with the use of raw and non-linguistic features such as Mel-Frequency Cepstral Coefficients (MFCCs), Linear Prediction Cepstral Coefficients (LPCCs)\cite{koolagudi2012emotion}, energy and pitch, and building classifiers to estimate the speaker's traits from these features. Other work in the literature focused on improving utterance modeling in low dimensional vectors such as i-vector \cite{li2015improved,martinez2011language,senoussaoui2010vector}, x-vector \cite{snyder2018x}, and d-vector \cite{variani2014deep}. These utterance modeling techniques showed significant enhancements in speaker verification and diarization tasks, hence researchers started to use them in speech classification tasks. The aforementioned features will be discussed in more detail later in this paper. other various classifiers were also explored: Gaussian Mixture Models (GMM), Hidden Markov Models (HMM), Support Vector Machines (SVM), Neural Networks, and other hybrid classifiers or ensembles \cite{klaylat2018emotion}.

\subsection{Arabic Speaker Profiling}
A lack of available Arabic speech corpora for different speech recognition tasks, and with only a few efforts reported in the attempt of Arabic speech recognition in comparison with other global languages, alongside the fact that Arabic is a very phonetically rich language and significantly differs between dialects \cite{elmahdy2011challenges,ali2017speech}. All the mentioned facts acclimatize the challenge of speech recognition and detection tasks for the Arabic language. There have been some attempts for speech classification tasks either by building labeled speech corpora or by implementing and proposing classifiers for gender, emotion, dialect, or other detection tasks \cite{klaylat2018emotion,QCRI,akbacak2011effective}. Examples of open-source data for Arabic Speech are Aljazeera that can be used for ASR, Qatar Computation, and Research Foundation Dialect Corpus. Other speech corpora can be acquired such as King Saud University speech corpus and emotion corpus.

The natural language processing (NLP) community has aggregated dialectal Arabic into five regional language groups: Egyptian (EGY), North African (NOR), Gulf or Arabian Peninsula (GLF), Levantine (LAV), and Modern Standard Arabic (MSA) \cite{QCRI}. The task of Arabic dialect detection solely from speech has been investigated by utilizing multiple approaches. Authors in \cite{QCRI} used SVM classifiers on top of i-vector-based utterances for the task of dialect detection. While other work in literature followed the approach of language identification \cite{QCRI,martinez2011language}. However, there have not been any reported attempts for Arabic dialect identification using multitask learning.

There are several categorizations for emotion in NLP. The most famous is Ekman's categorization \cite{ekman1997face}, which has the following categories: anger, disgust, fear, happiness, sadness, and surprise. However, in this paper, the emotions are categorized following \cite{meftah2018evaluation}'s categorization: anger, surprise, happiness, sadness, questioning, and neutral. Emotion detection from Arabic speech has been explored in previous work as in\cite{klaylat2018emotion,chen2018multi}, where authors used multiple well-known classification methods and ensembles on raw features such as MFCC and pitch.
Gender detection is an easier task in comparison with emotion and dialect detection in most languages including Arabic. The work in the literature performed well by building simple classifiers on raw features as in \cite{klaylat2018emotion}.

\subsection{Multi-task Speaker Profiling}
\label{ssec:Multi-task}
Multi-task learning is generally applied by sharing some hidden layers between all tasks while keeping several task-specific hidden and output layers, each with a loss function task\cite{zhang2017survey,zhang2018overview}.
Researchers started scratching the surface in exploring multi-task learning for speech recognition tasks after its successful utilization in text-based classification tasks. It has first been applied to enhance the robustness and phoneme recognition for ASR systems \cite{shinohara2016adversarial}. However, this paradigm application has been extended to speech classification tasks. In \cite{poorjam2014multitask}, authors used multi-task learning for estimating age, height, weight, and smoking habits of speakers, using utterance modeling techniques and other classifiers. Researchers in \cite{chen2018multi} proposed a multi-task learning approach to simultaneously estimate speaker identity and the other two traits of the speaker, specifically, emotion and gender from Arabic speech.

This paper proposes the use of a multi-task learning paradigm to simultaneously estimate the gender, dialect, and emotion of the speaker, which slightly outperformed single task classifiers on the three aforementioned tasks. All the datasets and pre-trained models assembled, edited, and used in this work are available to the public (either as open source or can be acquired). Also, various utterance modeling techniques, features, and network architectures are explored throughout the experiments and are reported in this paper. Finally, the code used to pre-process and prepare the data is available to future researchers.

\section{Datasets}
\label{sec:dataset}
This section briefly describes the Arabic Speech datasets collected from the literature. The datasets were collected for emotion, dialect, and gender recognition tasks. It is worth mentioning that for this research, only datasets that are either open-source or can be acquired by the public were used.\\

\textbf{Qatar Computing Research Institute (QCRI) Arabic dialect detection data:} The training corpus was collected from the Broadcast News domain in four Arabic dialects: Egyptian, North African, Gulf, Levant, and Modern Standard Arabic (MSA)\cite{QCRI}. The recordings were segmented to avoid speaker overlap, and any non-speech parts such as music and background noise were removed. The speech corpus consists of 15,000 utterances, with around 300 utterances for each dialect, and another 1,500 utterances as a development set.\\

\textbf{King Saud University Emotions (KSUEmotion) Corpus:} This speech corpus contains 16 different sentences spoken with 6 distinct emotions: Neutral, happy, sad, angry, surprised, and questioning\cite{meftah2018evaluation}. These 16 sentences are written in MSA and spoken by 23 different speakers from Syria, Yemen, and Saudi Arabia. 10 of these speakers are male while the other 13 are female, which makes this data suitable for both emotion detection and gender detection tasks. This dataset has 3,280 utterances divided equally between genders, while each emotion category has 650 utterances except for questioning and anger having 320 utterances each.\\

\textbf{Arabic Natural Audio Dataset (ANAD):} This Arabic speech corpus was developed to recognize 3 types of emotions from Arabic speech: Happy, angry, and surprised\cite{klaylatarabic}. Eight videos of live calls from online Arabic talk shows were downloaded and pre-processed by dividing the calls into turns: callers and receivers, and then silences, laughs, and noisy segments were removed. Finally, the utterances were labeled into happy, angry, and sad by 18 different Arabic listeners. However, while conducting the experiments, it was found that the duration of each segment was not long enough. To solve this, each segment was changed to have at least 3 seconds, and any silences were trimmed. The corpus after the new segmentation consists of 475 recordings.\\

\textbf{Spoken Arabic Regional Archive (SARA):} The contributors of this dataset collected from episodes and films published on YouTube played by native speakers with three different Arabic dialects and accents: Egyptian, the Arabian Peninsula dialect and Levantine dialect\cite{ziedan2016unified}. The dataset samples vary in length from 3 to 7 seconds in order to verify the minimum time required to detect the speaker's dialect. This corpus consists of 3,480 voice recordings divided evenly between the three dialects.\\

\textbf{Multi Dialect Arabic Speech (MDAS):} This corpus was designed to recognize three dialects along with Modern Standard Arabic (MSA). These dialects are Levant, Gulf, and Egyptian \cite{almeman2013multi}. The recordings were conducted with the consent of 52 participants. The recordings' duration totaled around 32 hours. After the segmentation stage, 67,132 speech files were obtained.\\

\textbf{King Saud University (KSU):} King Saud University Arabic Speech Database was developed by Speech Group (SG) at King Saud University and consists of 590 hours of recorded Arabic speech from 269 males and females speakers\cite{alsulaiman2013ksu}. The corpus was designed principally for speaker recognition research and other applications. However, in this research, it is used for gender detection. Only 49 hours were taken from the original corpus, which accounts for 9,050 utterances spoken by 350 different speakers.\\


All these datasets were aggregated together and then divided into train, development, and test sets while obeying to a couple of restrictions: (i) there should be no speaker overlap between the train, development, and test sets, (ii) each class of each dataset is almost divided between train(80\%), development(10\%) and test(10\%) sets. Following these rules implies that the entire dataset will be divided with the same percentages.
When abiding with these restrictions, it is almost impossible to get a perfect split because each speaker has several utterances labeled with multiple classes. Choosing the set for a speaker could be optimal for one class, but not the other. A dataset split using this format can be reduced into an integer programming problem.
Table \ref{tab:dataset_summary} shows the distribution of each class in each set.
\begin{table*}[tb]
\begin{center}
\small
\label{tab:dataset_summary}
\caption{Distribution of datasets}
\begin{tabular}{|p{2.2cm}|p{0.6cm}p{0.6cm}|
                p{0.6cm}p{0.6cm}p{0.6cm}p{0.6cm}p{0.6cm}|
                p{0.6cm}p{0.6cm}p{0.6cm}p{0.6cm}p{0.6cm}p{0.6cm}|
                p{0.7cm}|}
 \hline Dataset & \multicolumn{2}{c|}{Gender} &
        \multicolumn{5}{c|}{Dialect} &
        \multicolumn{6}{c|}{Emotion} &
        Total \\ \hline
Labels & M & F &  
MSA & LEV & GUL & NOR & EGY &          
NEU & SAD & HAP & SUR & ANG & QUES &        
TOT \\

\hline
    \multicolumn{14}{c|}{Train Set} \\
    \hline
    
    KSU & 1360 & 1357 &               
     0 & 0 & 0 & 0 & 0 &          
     544 & 542 & 544 & 544 & 288 & 255 &         
     5434 \\                
     
    ANAD & 0 & 0 &            
    0 & 0 & 0 & 0 & 0 &          
    0 & 0 & 122 & 33 & 210 & 0 &        
     365 \\                
     
    QCRI & 0  & 0  &          
    1998 & 2807 & 2586 & 2885 & 2860 &          
    0 & 0 & 0 & 0 & 0 & 0 &        
     13136 \\                
    
    SARA & 1691  & 1342  &          
    0 & 949 & 961 & 0 & 1123 &          
    0 & 0 & 0 & 0 & 0 &  0 &       
     6066 \\                
    
    MDAS & 0  & 0  &          
    12910 & 7746 & 12910 & 0 & 20655 &          
    0 & 0 & 0 & 0 & 0 & 0 &        
     54221 \\                  
    
    KING & 3887  & 3374  &          
    0 & 0 & 0 & 0 & 0 &          
    0 & 0 & 0 & 0 & 0 &  0 &       
      7261 \\                    
    \hline
    Total & 6938  & 6073  &         
     14908 & 11502 & 16457 & 2885 & 24638 &          
     544 & 542 & 666 & 577 & 498 & 255 &        
    86483  \\                    
      
    \hline

    \multicolumn{14}{c|}{Dev Set} \\
    \hline
    
KSU & 200 & 80 &               
     0 & 0 & 0 & 0 & 0 &          
     56 & 56 & 56 & 56 & 24 & 32 &        
     560 \\                
     
    ANAD & 0 & 0 &            
    0 & 0 & 0 & 0 & 0 &          
    0 & 0 & 19 & 4 & 32 & 0 &        
     55 \\                
     
    QCRI & 0  & 0  &          
    221 & 311 & 287 & 320 & 317 &          
    0 & 0 & 0 & 0 & 0 & 0 &        
     1456 \\                
    
    SARA & 224  & 204  &          
    0 & 146 & 125 & 0 & 157 &          
    0 & 0 & 0 & 0 & 0 &  0 &       
     856 \\                
    
    MDAS & 0  & 0  &          
    1291 & 1291 & 1290 & 0 & 2582 &          
    0 & 0 & 0 & 0 & 0 & 0  &       
     6454 \\                  
    
    KING & 482  & 410  &          
    0 & 0 & 0 & 0 & 0 &          
    0 & 0 & 0 & 0 & 0 & 0  &       
      892 \\                    
    \hline
    Total & 906  & 694  &         
     1512 & 1748 & 1702 & 320 & 3056 &          
     56 & 56 & 75 & 60 & 56 & 32 &        
    10273  \\                    
      
    \hline
    
    \multicolumn{14}{c|}{Test Set} \\
    \hline
    
    KSU & 79 & 200 &               
     0 & 0 & 0 & 0 & 0 &          
     56 & 55 & 56 & 56 & 24 & 32  &       
     558 \\                
     
    ANAD & 0 & 0 &            
    0 & 0 & 0 & 0 & 0 &          
    0 & 0 & 19 & 4 & 32 & 0 &        
     55 \\                
     
    QCRI & 0  & 0  &          
    283 & 348 & 265 & 355 & 315 &          
    0 & 0 & 0 & 0 & 0 & 0 &        
     1566 \\                
    
    SARA & 187  & 191  &          
    0 & 149 & 144 & 0 & 85 &          
    0 & 0 & 0 & 0 & 0 & 0 &        
     756 \\                
    
    MDAS & 0  & 0  &          
    1291 & 1291 & 1291 & 0 & 2582 &          
    0 & 0 & 0 & 0 & 0 & 0 &        
     6455 \\                  
    
    KING & 478  & 419  &          
    0 & 0 & 0 & 0 & 0 &          
    0 & 0 & 0 & 0 & 0 & 0 &        
      897 \\                    
    \hline
    Total & 744  & 810  &         
     1574 & 1788 & 1700 & 355 & 2982 &          
     56 & 55 & 75 & 60 & 56 & 32 &        
    10287  \\                    
      
    \hline

\end{tabular}
\end{center}
\end{table*}

\section{Feature Extraction}
\label{sec:feature_extraction}
This section briefly describes the features used to train the network. Five types of features are used in this paper: MFCC, MEL, i-vector, d-vector, and x-vector. They are briefly described below. It is worth mentioning that i-vector, d-vector, and x-vector extracted or trained during the experiments were either trained on data that is available to the public or extracted from available pre-trained models. Regarding the specification of the speech files, they were converted to the WAV file type, with a 16k sampling rate, mono-channel, and 16-bit precision. 

\subsection{Mel-spectrogram (Mel) Extraction}
Mel-spectrogram and MFCC are common methods for feature extraction from speech. A speech signal is first segmented into frames, each with a length between 20 to 40 ms to get a stationary signal. The Discrete Fourier Transform (DFT) is calculated for each frame in order to calculate the power density. Then, the frequency response calculated from DFT is passed through triangular band-pass filters to smoothen the energy, which is then finally converted to the Mel-scale. In this paper, parameter selection follows \cite{snyder2018x} where frames are 25 ms each. FFT size is 1024, and the number of filters is 24.


\subsection{Mel-Frequency Cepstral Coefficients (MFCC) Extraction}
Previously, the steps to compute filter banks energies in a log-spaced frequency domain were discussed, which are known as Mel-spectrogram features. These steps were originally motivated by the way humans perceive audio. 
An additional step of passing the computed log-filter banks energies is through a Discrete Cosine Transform (DCT). Using DCT will de-correlate the filter banks energies and transform the frequency domain into time-like domain features. A $L=14$ mel-scaled cepstral coefficients were calculated. The result coefficients are called Mel-Frequency Cepstral Coefficients (MFCC).

\subsection{I-Vector Extraction}
The need to represent speech utterances in a low dimensional, fixed-length vector regardless of the utterance length led to the emergence of i-vector-based frameworks\cite{li2015improved,martinez2011language,senoussaoui2010vector}
. I-vector captures the long-term characteristics of audio, such as the speaker's characteristics or recording device. Originally, this representation was developed for speaker recognition and verification, but it is now used in many areas of speech processing. Also, it has been widely used in the classification tasks in these papers, including dialect and emotion detection. In this paper, i-vectors of dimension 400 were trained on the dataset mentioned in the previous section. I-vector is considered to be the first generation of low dimensional fixed-length vector representations of speech, as it is based on the factor analysis on Gaussian Mixture Model (GMM) mean supervectors and the Non-negative Factor Analysis (NFA) framework, which is
based on a constrained factor analysis on GMM weights.

\subsection{D-Vector Extraction}
The d-vector features were extracted from a pre-trained network that had been trained on a text-independent speaker verification task using a large amount of data\cite{jia2018transfer,variani2014deep}.
The network input is 40-channel log-Mel spectrograms of a WAV of arbitrary length, passed to a stack of 3 LSTM layers of 768 cells, each followed by a projection to 256 dimensions. The final embedding is created by L2-normalizing the output of the top layer at the final frame. The output is an embedding vector of length 256 known as a d-vector.
\subsection{X-Vector Extraction}
X-vectors are fixed-dimensional embeddings extracted from a network trained on a speaker verification task\cite{snyder2018x}. Embeddings of this type need substantial amounts of data, which called for the need to augment the data by adding noise and reverberations as an inexpensive way to multiply the training data. Furthermore, it was found that x-vector benefited more from the augmentation techniques compared to i-vector and performed better in speaker verification tasks. In this paper, x-vectors were extracted from a pre-trained model on large data.

\bigbreak

All these features were pre-computed and stored into separate files, each containing the feature type for audio files in all datasets. This step was performed for two reasons: (i) to make the training process fast and efficient, (ii) to remove any indeterministic nature that could exhibit some of these extraction methods. We refrain from disclosing those files due to license agreements for some of the datasets which forbid redistribution.

\section{Method}
\label{sec:method}
This section describes the multi-task learning networks implemented, trained, and their combinations. It also describes the tuned hyper-parameters and the training setup of the experiments.

\subsection{Network Architecture}
There are many approaches to build a network that tackles the problem of multi-task speech classification. In our research, machine learning blocks that can be utilized in combination with specific input features to formulate an end-to-end prediction model are identified and briefly described.
Figure \ref{fig:sparta_method} illustrates the building blocks of the SPARTA network. The illustration does not represent a network, but a framework on which components can be connected to create a network. These building blocks are grouped into 3 main sections:

\begin{figure*}[h]
    \centering
    \includegraphics[width=1.0 \textwidth]{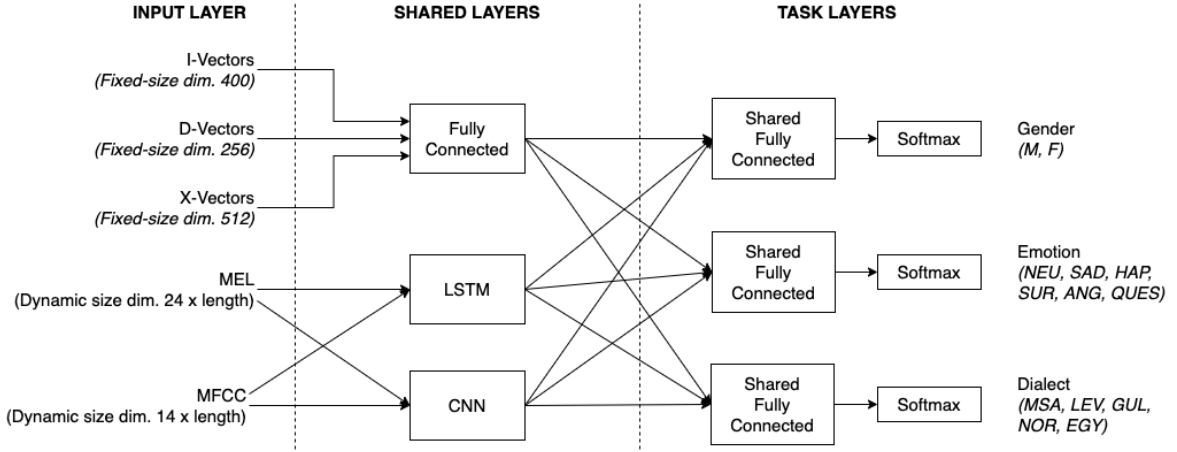}
    \caption{Building blocks of SPARTA framework}
    \label{fig:sparta_method}
\end{figure*}

\begin{enumerate}
    \item \textbf{Input Layer:} This layer was designed to extract feature representations from an utterance signal. These features, as described in the previous section, are either fixed-size learned feature extractions using methods such as i-vector, d-vector, and x-vector or dynamic-size frequency cepstrum representations such as MFCC and MEL.
    Given that fixed-size features learn different representations, concatenating these vectors could enrich the feature space. All possible permutations of vectors are generated: id-vector, ix-vector, dx-vector, and idx-vector.
    However, since MFCC is a compressed version of MEL, they were not concatenated, as having both would produce redundant information.
    
    \item \textbf{Shared Layers:} These layers consume the output of the previous layer and transform it into an intermediate representation that captures more information to help classify the selected tasks. The type of shared layer chosen is dependent on the selected features. Fully connected layers are chosen to consume dense and fixed input representations (i, d, x vectors, and their combinations), while either convolutional layers or LSTM are chosen for variable-size inputs (MFCC and MEL). Certain combinations were omitted, such as consuming a fixed-size vector with an LSTM network which is designed to consume sequential variable length input. It is worth noting that the weights of these layers are tweaked by the error propagated from all tasks.
    
    \item \textbf{Task Layers:} These layers are composed of several sub-networks where each sub-network consumes output from the shared layers and computes predictions for its assigned task. Each sub-network is built from a configurable number of fully connected layers, and ending with a softmax layer. Cross entropy loss is used across all tasks. Note that parameters are not shared between tasks, hence the back-propagation algorithm only tweaks one sub-network per batch. When performing single-task learning, only one prediction layer is active while the other two are disabled.
\end{enumerate}

There are eleven possible variations available to connect the aforementioned blocks to build a multi-task speech classification model. All these models were built and trained with the dataset described in section \ref{sec:dataset} in both MLT and STL format.

\subsection{Experimental Setup}
As mentioned in the previous section, three network architectures were implemented followed by task-specific networks. The hyper-parameters of the networks were explored to obtain the best possible combination of parameters that results in the best accuracy and F1 score. Table \ref{tab:hyperparameters} presents each hyper-parameter and their ranges.
\begin{table*}[h]
\caption{Networks hyper-parameters}
\label{tab:hyperparameters}
\small
\centering
\begin{tabular}{ccc|ccc}
Network                          & Hyper-parameter   & Range                                                                 & Network                                                                          & Hyper-parameter  & Range                                                                  \\ \hline
\multirow{6}{*}{CNN}             & Features          & MFCC, MEL                                                             & \multirow{6}{*}{LSTM}                                                            & Features         & MFCC, MEL                                                              \\
                                 & Number of filters & {[}32, 64{]}                                                           &                                                                                  & Direction        & \begin{tabular}[c]{@{}l@{}}Uni/bi- directional\end{tabular}          \\
                                 & Filter size       & {[}3, 4, 5{]}                                                           &                                                                                  & Hidden nodes     & \begin{tabular}[c]{@{}l@{}}{[}16, 32, 64, 128{]}\end{tabular}       \\
                                 & Number of layers  & {[}1, 2, 4{]}                                                           &                                                                                  & Number of layers & {[}1,2,4{]}                                                            \\
                                 & Activation        & {[}relu, tanh{]}                                                      &                                                                                  & Activation       & {[}relu, tanh{]}                                                       \\
                                 & Dropout           & {[}0, 0.1, 0.2, 0.3{]}                                                &                                                                                  & Dropout          & \begin{tabular}[c]{@{}l@{}}{[}0, 0.1, 0.2, 0.3{]}\end{tabular}      \\ \hline
\multirow{5}{*}{Fully Connected} & Features          & \begin{tabular}[c]{@{}l@{}}i/d/x/id/ix/dx/idx-vector\end{tabular}  & \multirow{5}{*}{\begin{tabular}[c]{@{}l@{}}Task Specific \\ layers\end{tabular}} & Hidden nodes     & \begin{tabular}[c]{@{}l@{}}{[}16, 32, 64, 128{]}\end{tabular}       \\
                                 & Hidden nodes      & \begin{tabular}[c]{@{}l@{}}{[}16, 32, 64, 128, 256{]}\end{tabular} &                                                                                  & Number of layers & {[}1,2{]}                                                              \\
                                 & Number of layers  & {[}1,2,4{]}                                                           &                                                                                  & Activation       & \begin{tabular}[c]{@{}l@{}}{[}relu, tanh, sigmoid{]}\end{tabular}   \\
                                 & Activation        & \begin{tabular}[c]{@{}l@{}}{[}relu, tanh, sigmoid{]}\end{tabular}  &                                                                                  & Dropout          & \begin{tabular}[c]{@{}l@{}}{[}0, 0.1, 0.2, 0.3, 0.5{]}\end{tabular} \\
                                 & Dropout           & {[}0, 0.1, 0.2, 0.3{]}                                                &                                                                                  &                  &                                                                        \\ \hline
\end{tabular}
\end{table*}

In addition to exploring the network hyper-parameters, different training parameters and setups were investigated, including the optimizer, the learning rate and its decay, and the number of epochs. Furthermore, in order to balance the number of utterances for each task, a multiplication factor for the utterances of a selected task was added to ensure the balance of the datasets for all the tasks. The training setup has two variations: (i) train one task at a time wherein each epoch the model finishes training on one task then moves to the next, or (ii) shuffled where each batch is selected at random from any task.

A combination of manual and grid search algorithm \cite{lecun1998neural} was applied to explore all tuning parameters. The model was evaluated against the dev set after each epoch in each training process. The best epoch's weights were stored, and the final scores were computed by evaluating the best model on the test set as seen in table \ref{tab:dataset_summary}.

Finally, it is worth mentioning that the networks described in this section were implemented using the TensorFlow framework \cite{abadi2016tensorflow} and trained on GPUs of type Tesla T4.

\section{Results and Discussion}
\label{sec:results_and_discussion}
To evaluate the multi-task learning paradigm against the conventional single-task learning for the aforementioned tasks, several experiments were conducted on the previously described networks with different features, while varying the hyper-parameters of the networks in an attempt to obtain the best possible results. Also, to create benchmarks for the used datasets, the results of each dataset separately are reported in the section.

\subsection{Network Architecture Results}
The best results reported for each network and features in terms of F1 scores are presented in table \ref{tab:network_results}. From this table and throughout the experiments, it is evident that utterance modeling techniques such as i-vectors and d-vector outperform raw features (MFCC and MEL), especially in dialect and emotion detection tasks. This can be explained by the fact that these techniques take into consideration phonetic and lexical features with only a limited amount of data, unlike raw features that require far more hours of speech to capture such features on their own. Also, the performance resulting from the concatenation of i-vector, d-vector, and x-vector surpasses their individual performances, as each one of these captures different characteristics since they are trained on different and larger datasets. It is noteworthy that the performance of d-vector on its own is relatively poor. However, when concatenated with i-vector and x-vector the model's performance was boosted as it contains a different information type than the others and has been trained on different kinds of data. Furthermore, in the case of pre-trained vectors, multi-task learning outperformed single-task learning in each task. However, the difference was more apparent in the emotion and dialect detection tasks. To further indicate the importance of utterance modeling techniques, in comparison to MFCC and MEL, utterance modeling techniques achieved better results in the gender detection task. This can be attributed to the fact that these techniques were originally designed for speaker verification tasks, hence they can recognize the gender of the speaker better.

On the other hand, CNN and LSTM networks require more data to surpass the performance of fully connected networks with utterance modeling techniques. Also, in the case of CNN adn LSTM, multi-task learning slightly enhanced the performance of the model when compared to single-task learning, which can be attributed to the fact that the network had more data and more shared layers, which positively affected that network's performance. However, LSTM and CNN performed less in comparison to other networks, due to the networks degrading ability to capture learning in proportion to the length of the sequence. 


\begin{table*}[h]
\centering
\small
\caption{Results (F1 scores) for Multi-task and single-task for each task}
\label{tab:network_results}
\begin{tabular}{cccccccccc}
\multirow{2}{*}{Network} &
  \multirow{2}{*}{Features} &
  \multicolumn{2}{c}{Gender} &
  \multicolumn{2}{c}{Emotion} &
  \multicolumn{2}{c}{Dialect} &
  \multicolumn{2}{l}{Overall} \\ \cline{3-10} 
                                 &            & STL   & MTL            & STL   & MTL            & STL   & MTL            & STL   & MTL            \\ \hline
\multirow{7}{*}{Fully connected} & d-vector   & 99.10 & 98.58          & 42.60 & 55             & 36.75 & 37.33          & 59.48 & 63.63          \\
                                 & i-vector   & 92.59 & 93.88          & 55.9  & 58.35          & 46.61 & 46.67          & 65.03 & 66.3           \\
                                 & x-vector   & 98.90 & 99.61          & 54.13 & 54.48          & 48.61 & 48.77          & 67.21 & 67.62          \\
                                 & id-vector  & 97.10 & 96.39          & 62.88 & 66.47          & 51.14 & 53.44          & 70.34 & 72.1           \\
                                 & ix-vector  & \textbf{99.42} & 99.48          & 62.26 & 61.98          & 52.33 & 52.59          & 71.41 & 71.35          \\
                                 & dx-vector  & 98.71 & \textbf{99.81} & 57.14 & 64.24          & 52.35 & 51.07          & 69.4  & 71.71          \\
                                 & idx-vector & 99.10 & 99.41          & \textbf{66.53} & \textbf{70.16} & \textbf{56.05} & \textbf{58.44} & \textbf{73.89} & \textbf{76.01} \\ \hline
\multirow{2}{*}{CNN}             & MFCC       & 96.13 & 95.54          & 52.91 & 51.95          & 21.19 & 26.7           & 56.98 & 58.08          \\
                                 & MEL        & 97.16 & 95.54          & 51.18 & 51.25          & 22.74 & 27.89          & 57.02 & 58.22          \\ \hline
\multirow{2}{*}{LSTM}            & MFCC       & 96.39 & 95.49          & 54.9  & 57.04          & 26.8  & 29.07          & 59.3  &   60.53        \\
                                 & MEL        & 94.96 & 95.12          & 50.4  & 57.05          & 28    & 27.94           & 57.78 &  60.2           \\ \cline{2-10} 
\end{tabular}
\end{table*}


\subsection{Data Specific Results}
Table \ref{tab:dataset_results} illustrates the F1 score obtained for each dataset separately on single-task learning and multi-task learning (all datasets combined). This table sets a benchmark for all datasets used in this work. It can be noticed that the results of multi-task learning almost persistently outperform the single-task learning for each dataset, especially when using pre-trained vectors for utterances. The effect of multi-task learning enhancement is especially evident in the emotion and dialect detection tasks. However, on the gender tasks, the results are close between the single task learning and multi-task learning. 
The work in \cite{chen2018multi} used the KSUEmtotion dataset for emotion and gender recognition; the accuracy recorded was 79.3\% for the task of emotion recognition and 98.7\% on gender detection. The best experiment of the proposed model outperformed these results as shown in table \ref{tab:dataset_results}.


\begin{table*}[h]
\centering
\small
\caption{Results (F1 scores) for Multi-task and single-task for each each dataset (G: Gender, E: Emotion, D: Dialect)}
\label{tab:dataset_results}
\begin{tabular}{cccccccccc}
\multirow{3}{*}{Model}                               & \multirow{3}{*}{Features} & \multicolumn{8}{c}{Dataset}                                                                                                           \\ \cline{3-10} 
                                                     &                           & \multicolumn{2}{c}{KSUEmotion G} & \multicolumn{2}{c}{KSUEmotion E} & \multicolumn{2}{c}{ANAD E}     & \multicolumn{2}{c}{SARA G}     \\ \cline{3-10} 
                                                     &                           & ST              & MT             & ST             & MT              & ST            & MT             & ST            & MT             \\ \hline
\multirow{2}{*}{CNN}                                 & MFCC                      & 96.5            & 97.4           & 50.7           & 50.35           & 56.3          & 54.93          & 87.8          & 85.8           \\
                                                     & MEL                       & 92.9            & 94.1           & 50.51          & 50.13           & 51.11         & 50.81          & 83.8          & 87.1           \\ \hline
\multirow{2}{*}{LSTM}                                & MFCC                      & 80.1            & 98.25          & 51.9           & 55.53           & 32.3          & 44.64          & 85.2          & 87.39          \\
                                                     & MEL                       & 81.4            & 95.32          & 46.1           & 54.79           & 34.2          & 44.34          & 83            & 84.96          \\ \hline
\multirow{7}{*}{Fully Connected}                     & d-vector                  & 100             & 100            & 48.1           & 52.4            & 72.1          & 73.6           & 94.4          & 96.8           \\
                                                     & i-vector                  & 100             & 100            & 62.6           & 66.7            & 88.1          & 85.36          & 79.3          & 79.5           \\
                                                     & x-vector                  & 100             & 100            & 66.32          & 69.54           & 88.8          & 84.34          & 94.9          & 98.4           \\
                                                     & id-vector                 & 100             & 100            & 61.56          & 64.36           & 87.1          & 85.46          & 86.3          & 87.5           \\
                                                     & ix-vector                 & 100             & 100            & \textbf{73.4}  & 75.4            & 90.3          & 91             & 98.4          & 97.8           \\
                                                     & dx-vector                 & 100             & 100            & 67             & 70.3            & 85.9          & 83.87          & 95.4          & \textbf{99.4}  \\
                                                     & idx-vector                & \textbf{100}    & \textbf{100}   & 72.34          & \textbf{80.35}  & \textbf{91.8} & \textbf{91.6}  & \textbf{99.4} & 97.6           \\ \hline
\multicolumn{1}{l}{}                                 & \multicolumn{1}{l}{}      & \multicolumn{2}{c}{SARA D}       & \multicolumn{2}{c}{KSU G}        & \multicolumn{2}{c}{MDAS D}     & \multicolumn{2}{c}{QCRI D}     \\ \cline{3-10} 
\multicolumn{1}{l}{}                                 & \multicolumn{1}{l}{}      & ST              & MT             & ST             & MT              & ST            & MT             & ST            & MT             \\ \hline
\multicolumn{1}{l}{\multirow{2}{*}{CNN}}             & \multicolumn{1}{l}{MFCC}  & 26.6            & 25.9           & 98.9           & 98.8            & 24.2          & 23.9           & 25.8          & 25.2           \\
\multicolumn{1}{l}{}                                 & \multicolumn{1}{l}{MEL}   & 26.1            & 24.82          & 99.1           & 99.2            & 22.9          & 25.7           & 25.4          & 23.8           \\ \hline
\multicolumn{1}{l}{\multirow{2}{*}{LSTM}}            & \multicolumn{1}{l}{MFCC}  & 30.4            & 22.1           & 98.7           & 97.87           & 43.4          & 27.6           & 30.6          & 27.5           \\
\multicolumn{1}{l}{}                                 & \multicolumn{1}{l}{MEL}   & 28.8            & 21.2           & 98.3           & 97.13           & 30.7          & 25.91          & 31.0          & 28.49          \\ \hline
\multicolumn{1}{l}{\multirow{7}{*}{Fully Connected}} & d-vector                  & 40.1            & 36.32          & \textbf{100}   & 99.6            & 40.7          & 38.36          & 34.7          & 29.1           \\
\multicolumn{1}{l}{}                                 & i-vector                  & 44.7            & 44.67          & 98.6           & 97.9            & 47.5          & 45.68          & 52.3          & 50.93          \\
\multicolumn{1}{l}{}                                 & x-vector                  & 38.3            & 37.93          & 99.8           & 100             & 56.7          & 53.7           & 38.5          & 37.5           \\
\multicolumn{1}{l}{}                                 & id-vector                 & 55.4            & 50.3           & 99.6           & 99.3            & 55.9          & 53.21          & 50.7          & 50.8           \\
\multicolumn{1}{l}{}                                 & ix-vector                 & 39.9            & 40.12          & 99.8           & \textbf{100}    & 60            & 56.87          & 49.6          & 47.3           \\
\multicolumn{1}{l}{}                                 & dx-vector                 & 31.2            & 30.45          & 99.8           & 99.8            & 55.9          & 54.3           & 39.4          & 36.8           \\
\multicolumn{1}{l}{}                                 & idx-vector                & \textbf{50.4}   & \textbf{49.8}  & 99.8           & 99.88           & \textbf{60.6} & \textbf{70.67} & \textbf{53.6} & \textbf{54.46} \\ \cline{2-10} 
\end{tabular}
\end{table*}

\section{Conclusion and Future Work}
\label{sec:conclusion}
This paper presents a novel approach that utilizes multi-task learning for the tasks of gender, emotion, and dialect detection from Arabic speech. It also explores various pre-trained vectors, their concatenation, and raw features on different networks and implementations. Additionally, it introduces a new Arabic Speech benchmark collected from publicly available datasets with code to reproduce the features and pre-trained vectors.

Throughout the experiments conducted in this paper, the following can be concluded. First, pre-trained vectors such as i-vector, d-vector, and x-vector enhance any network's performance on detection tasks and should be further explored and improved on. Also, similar to text classification tasks, multi-task learning improves performance on speech classification and recognition tasks. Therefore, its utilization must be expanded further on speech-related tasks. Moreover, despite the significant shortage of publicly available corpora for Arabic speech, speech corpora that can be used for gender, emotion, and dialect detection tasks are assembled with benchmarks for each task and dataset.
Future work of this paper should include enhancing pre-trained vectors for utterance modeling by training them on larger Arabic corpus and include available Arabic ASR data. Furthermore, explore more architectures and features such as self-attention, TDNN, and filter banks.

\bibliographystyle{unsrt}
\bibliography{sparta}

\end{document}